%% file: main.tex
\documentclass[10pt,twocolumn,letterpaper]{article}

 \usepackage[pagenumbers]{cvpr} 

\usepackage[accsupp]{axessibility} 

\input{preamble}

\definecolor{cvprblue}{rgb}{0.21,0.49,0.74}
\usepackage[pagebackref,breaklinks,colorlinks,citecolor=cvprblue]{hyperref}
\def\paperID{7432} 
\def\confName{CVPR}
\def\confYear{2024}

\title{CN-RMA: Combined Network with Ray Marching Aggregation for 3D Indoor Object Detection from Multi-view Images}

\author{
    Guanlin Shen$^{1}$ \qquad
    Jingwei Huang$^{2}$ \qquad
    Zhihua Hu$^{3}$ \qquad
    Bin Wang$^{1}$\footnotemark[1] \\[3pt]
    $^{1}$School of Software, Tsinghua University, China \quad
    $^{2}$Tencent, China \\
    $^{3}$Nanjing University of Information Science and Technology, China 
}

\begin{document}
\maketitle
\footnotetext[1]{*Corresponding author, email address: wangbins@tsinghua.edu.cn. This work was supported by the National Natural Science Foundation of China under Grant 62072271.}

\input{sec/0_abstract}    
\input{sec/1_intro}
\input{sec/2_related_work}
\input{sec/3_method}

\input{sec/4_experiments}
\input{sec/5_conclusion}

{
    \small
    \bibliographystyle{ieeenat_fullname}
    \bibliography{main}
}

\input{sec/X_suppl}

\end{document}

%% file: preamble.tex
\usepackage{bbm}
\usepackage{newtxmath}
\usepackage[dvipsnames]{xcolor}


%% file: sec/0_abstract.tex
\begin{abstract}
This paper introduces CN-RMA, a novel approach for 3D indoor object detection from multi-view images.
We observe the key challenge as the ambiguity of image and 3D correspondence without explicit geometry to provide occlusion information.
To address this issue, CN-RMA leverages the synergy of 3D reconstruction networks and 3D object detection networks, where the reconstruction network provides a rough Truncated Signed Distance Function (TSDF) and guides image features to vote to 3D space correctly in an end-to-end manner. 
Specifically, we associate weights to sampled points of each ray through ray marching, representing the contribution of a pixel in an image to corresponding 3D locations. Such weights are determined by the predicted signed distances so that image features vote only to regions near the reconstructed surface.
Our method achieves state-of-the-art performance in 3D object detection from multi-view images, as measured by mAP@0.25 and mAP@0.5 on the ScanNet and ARKitScenes datasets. The code and models are released at \href{https://github.com/SerCharles/CN-RMA}{https://github.com/SerCharles/CN-RMA}.
\end{abstract}

%% file: sec/1_intro.tex
\section{Introduction}
\label{sec:intro}

\begin{figure*}[t]
  \centering
   \includegraphics[width=1.0\linewidth]{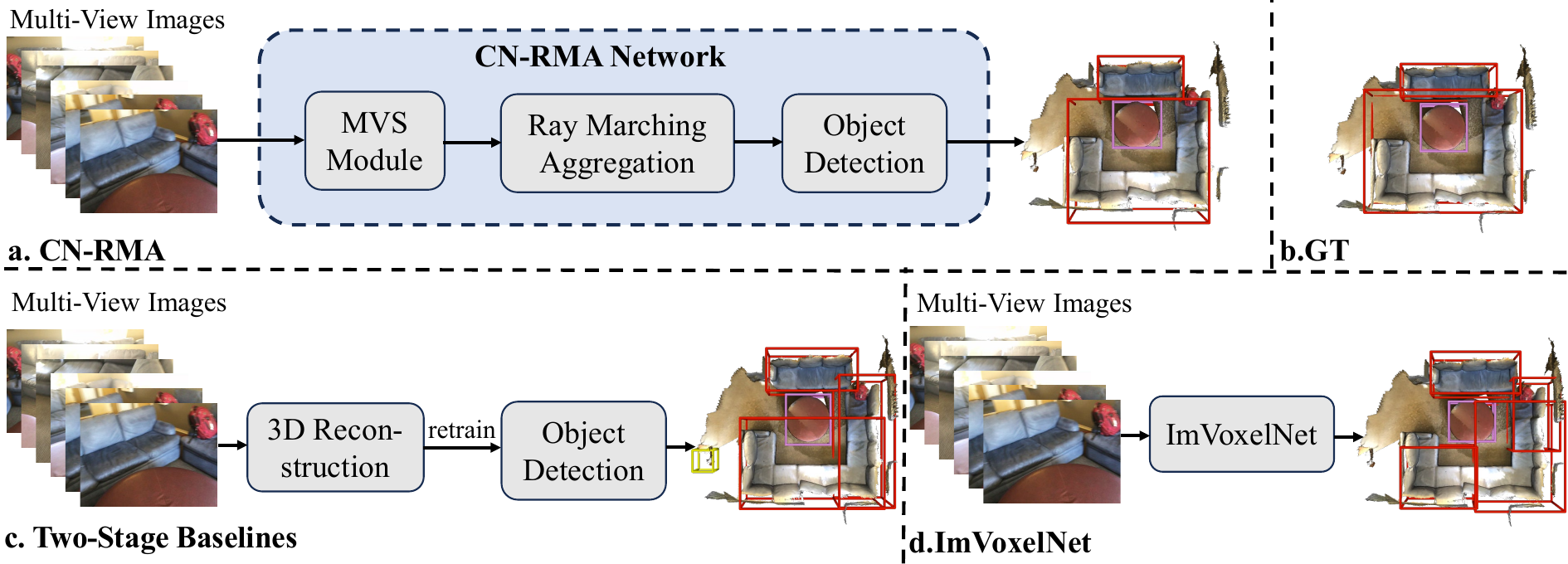}

   \caption{\textbf{The Comparison of Our CN-RMA, the two-stage method, and ImVoxelNet\cite{rukhovich2022imvoxelnet}.} Our CN-RMA is an end-to-end object detection method that incorporates an occlusion-aware 2D to 3D aggregation technique. In contrast, the two-stage method lacks end-to-end trainability, while ImVoxelNet employs a heuristic aggregation method that disregards occlusion considerations.}
   \label{fig:teaser}
\end{figure*}

 3D object detection from multi-view images is a fundamental problem in various fields, including robotics, autonomous driving, and augmented reality (AR). 
However, since explicit scene geometry is unavailable to detect occlusion, it is an ill-posed problem to identify correspondences between image regions and 3D locations. Therefore, image features can be wrongly projected to 3D, leading to inaccurate detection.
While occlusion is not a critical issue in open space and is ignored for autonomous driving scenarios~\cite{liu2022petr, wang2022detr3d}, it commonly exists among objects in complex environments like indoor scenes.

One straightforward solution for 3D object detection from multi-view images is to perform 3D scene reconstruction from multi-view images~\cite{murez2020atlas, sun2021neuralrecon, ren2023volrecon, wang2021neus, wang2022hf, choe2021volumefusion} followed by 3D object detection from reconstructed point clouds~\cite{zhou2018voxelnet, rukhovich2022fcaf3d, gwak2020generative}.  However, such a solution is not ideal due to the lack of connectivity between the two stages.
The first stage usually introduces noises and incompletion in the reconstructed 3D geometry with limited power of 3D reconstruction techniques,  and such geometry loss is not resolved in the second stage. Moreover, aggregated color signals to inaccurate geometry cannot fully exploit rich image features and further harm the performance.
An alternative method proposed by ImVoxelNet~\cite{rukhovich2022imvoxelnet} involves aggregating 2D features extracted from multi-view images into 3D voxel volumes through unprojection in an end-to-end manner. However, due to the lack of scene geometry information, this exploratory aggregation approach struggles to effectively address complex occlusion issues, leading to feature voting from images to unrelated 3D locations.

In this paper, we present CN-RMA, an end-to-end novel 3D object detection method from multi-view images that seamlessly combines the reconstruction and detection networks with occlusion-aware feature aggregation. Our network mainly consists of a Multi-View Stereo (MVS) module~\cite{murez2020atlas} and a novel occlusion-aware aggregation module followed by a 3D detection module~\cite{rukhovich2022fcaf3d}. In the MVS module, we aim to reconstruct the rough scene geometry. We extract 2D features from the input images and feed them into the reconstruction network to generate a rough Truncated Signed Distance Function (TSDF)~\cite{curless1996volumetric} as a 3D representation. Our key contribution is the occlusion-aware aggregation module called Ray Marching Aggregation (RMA), which leverages the reconstructed TSDF to detect occlusion based on ray marching. In comparison with conventional 3D detection methods that vote image features equally along rays to the 3D space, we associate different weights according to the signed distance values.
Specifically, RMA incorporates the idea of volume density given TSDF inspired by NeuS~\cite{wang2021neus} and accumulates transmittance through ray marching to calculate the weight of each point along a ray, effectively addressing the occlusion issues encountered in complex environments. Then, we can aggregate image features by weights in the 3D space aware of occlusion.
Finally, we extract points with aggregated features near the reconstructed surface and pass the point cloud to the 3D detection module for object detection.
Given the challenging task, we propose a pre-training and fine-tuning method to train the entire network, making the components cooperate to achieve the best performance.
Figure~\ref{fig:teaser} illustrates the comparison between our proposed CN-RMA, ImVoxelNet~\cite{rukhovich2022imvoxelnet} and the two-stage method.

We evaluate our method on the ScanNet~\cite{dai2017scannet} and ARKitScenes~\cite{dehghan2021arkitscenes} datasets to assess its performance and compare it with existing methods. Our approach outperforms other methods, achieving significant improvements in mAP@0.25 and mAP@0.5, including 3.2 and 3.0 in ScanNet, and 7.4 and 13.1 in ARKitScenes, respectively.

In summary, our contributions are three-fold:
\begin{itemize}
\item[$\bullet$]We establish a seamless connection between the multi-view 3D reconstruction network and 3D object detection network, enabling better exploitation of image features in 3D space for improved performance.
\item[$\bullet$]We propose an innovative occlusion-aware aggregation method, RMA, which leverages the reconstructed scene TSDF to address the complex occlusion issues.
\item[$\bullet$]We adopt a pretraining and finetuning scheme, and achieve the state-of-the-art performance for indoor 3D object detection from multi-view images.
\end{itemize}

%% file: sec/2_related_work.tex
\section{Related Work}
\label{sec:related}
\subsection{3D Object Detection from Multi-View Images}
3D object detection from multi-view images has been a hot topic in the vision community for many years. It aims to estimate the classes, poses, and sizes of objects from images. For outdoor scenes, a lot of methods project the features to Bird's Eye View (BEV) for the sake of memory and better performance~\cite{huang2021bevdet, li2022bevformer, wang2023frustumformer, lang2019pointpillars, wang2023dsvt}. However, the BEV representation is not suitable for 3D object detection in indoor scenes due to object stacking and occlusion. In recent years, several methods have tried to aggregate 2D features in 3D space~\cite{rukhovich2022imvoxelnet, wang2022detr3d, liu2022petr}. For instance, ImVoxelNet~\cite{rukhovich2022imvoxelnet} projects the 2D features from images into 3D space and aggregates the features with 3D CNN in the voxel form. DETR3D~\cite{wang2022detr3d} detects objects by generating random 3D object queries and linking 3D positions to images with camera transformation. PETR~\cite{liu2022petr} produces the 3D position-aware features by encoding the position information of 3D coordinates into image features. However, these exploratory aggregation methods have not taken full advantage of the scene geometry. ImGeoNet~\cite{tu2023imgeonet} introduces geometry implicit, while NeRF-Det~\cite{xu2023nerfdet} incorporates NeRF~\cite{mildenhall2021nerf}. However, these methods have not considered the occlusion during the feature aggregation process, leading to inaccurate detection results.

\subsection{Neural Implicit Reconstruction}
To recover the 3D geometry from multi-view images, neural implicit representations are often adopted, such as the Signed Distance Function (SDF) ~\cite{curless1996volumetric, izadi2011kinectfusion, murez2020atlas, sun2021neuralrecon, choe2021volumefusion}. The scene mesh can be obtained from the SDF using techniques like Marching Cubes~\cite{lorensen1998marching}. For instance, Atlas~\cite{murez2020atlas} predicts the Truncated Signed Distance Function (TSDF) of the scene with 3D CNN. And many subsequent methods have made improvements based on the Atlas network. NeuralRecon~\cite{sun2021neuralrecon} splits one complete scene into fragments and uses a Gated Recurrent Unit (GRU)~\cite{chung2014empirical} network to fuse the 3D features of the fragments to save time and memory. VolumeFusion~\cite{choe2021volumefusion} employs deep MVS~\cite{yao2018mvsnet} techniques to predict the TSDF. However, these improvements, while enhancing the effectiveness of 3D reconstruction, have made the network more complex and challenging to combine with other networks.

In recent years, NeRF~\cite{mildenhall2021nerf} based methods have utilized neural implicit fields in novel view synthesis and 3D reconstruction~\cite{yu2021pixelnerf, chen2021mvsnerf, wang2021ibrnet, long2022sparseneus}. 

For example, NeuS~\cite{wang2021neus} and VolSDF~\cite{yariv2021volume} incorporate the SDF into neural radiance fields by integrating it into the density function, bridging the gap between the SDF and the volume density of points along each ray. It shows the possibility of sampling and weighting points in 3D space with the Truncated Signed Distance Function (TSDF), which can be used to address the occlusion issues.

\subsection{3D Object Detection from Point Clouds}
3D object detection from point clouds is much more straightforward. According to the representation of point clouds, it can be divided into point cloud-based and voxel-based methods. For point cloud-based methods, the voting scheme introduced by VoteNet~\cite{qi2019deep} is broadly adopted~\cite{wang2021ibrnet, xie2021venet}, while PointNet++~\cite{qi2017pointnet++} is often used to extract the features of point clouds. However, voting-based object detection methods require additional data annotation on the point clouds, making them difficult to combine with reconstruction networks. Voxel-based methods usually convert point clouds into 3D voxels and utilize 3D CNN for voxel processing~\cite{zhou2018voxelnet, lang2019pointpillars}. However, dense volumetric representation and 3D CNN are memory-consuming. Thus, sparse convolution based on sparse voxels has been introduced to improve the performance of 3D object detection~\cite{gwak2020generative, rukhovich2022fcaf3d}.

Our approach focuses on enhancing the performance of 3D indoor object detection from multi-view images by integrating 3D scene reconstruction methods and 3D object detection methods from point clouds. To take full advantage of the scene geometry and handle occlusions, we designed an occlusion-aware aggregation method based on neural implicit representations.

%% file: sec/3_method.tex
\section{Method}
\label{sec:method}

\begin{figure*}[t]
  \centering
   \includegraphics[width=1.0\linewidth]{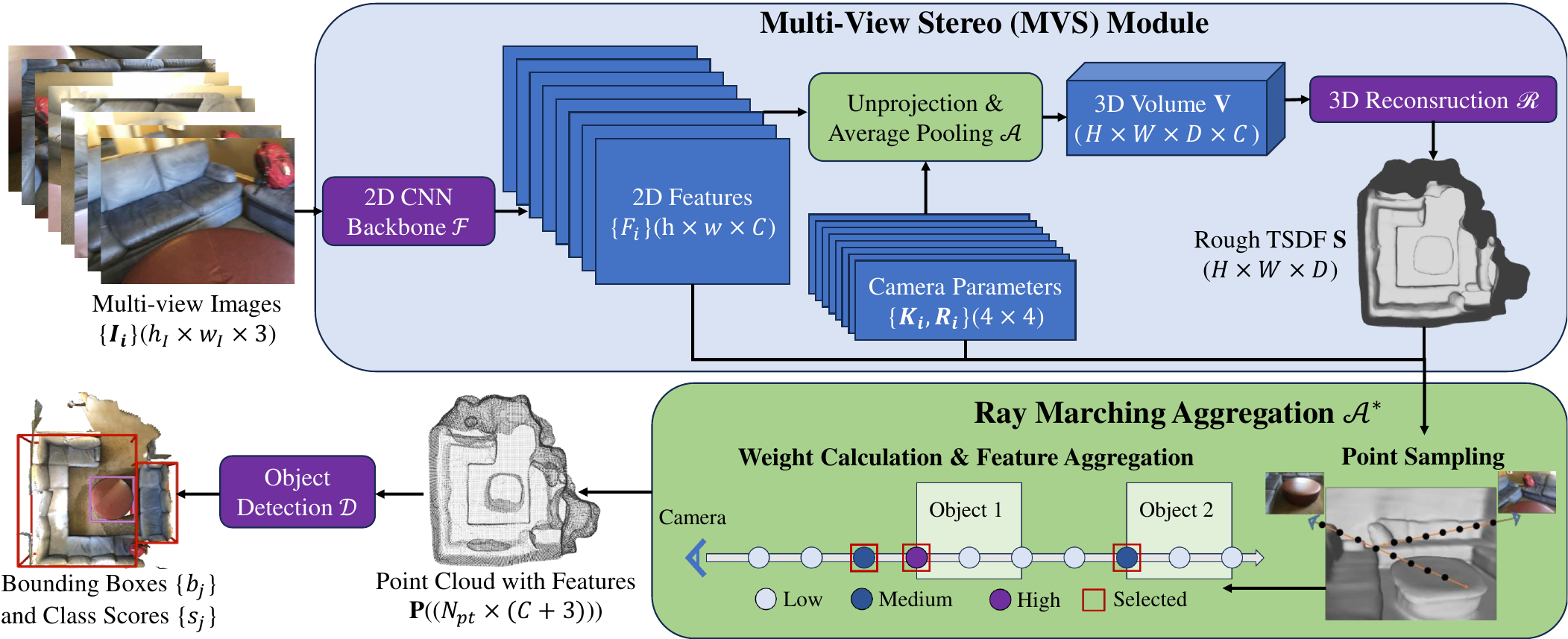}

   \caption{\textbf{The overall architecture of our CN-RMA method.} The purple blocks represent neural networks, while the green blocks represent modules without trainable neurons. Following Atlas~\cite{murez2020atlas}, the 2D CNN backbone $\mathcal{F}$ is a ResNet50-FPN network~\cite{lin2017feature}, and the 3D reconstruction network $\mathcal{R}$ is a 3D CNN network that features an encoder-decoder structure with skip connections with a $1\times1\times1$ convolutional head. Following FCAF3D~\cite{rukhovich2022fcaf3d}, the object detection network $\mathcal{D}$ is a sparse 3D convolutional network comprising a ResNet34 backbone~\cite{choy20194d} and a 4-layer decoder network.}
   \label{fig:overview}
\end{figure*}

\subsection{Problem Formulation}
We aim to achieve precise 3D object detection in a cluttered scene with complex occlusions using multi-view images and their corresponding camera parameters. To accomplish this, we propose a pipeline that combines a MVS reconstruction module and a 3D detection network through our occlusion-aware aggregation approach, as shown in Figure~\ref{fig:overview}. 

Given the input images $\{\textbf{I}_{i} \in R^{h_{I} \times w_{I} \times 3}\}$, along with corresponding camera intrinsics $\{\textbf{K}_{i} \in R^{3 \times 3}\}$ and extrinsics $\{\textbf{R}_{i} \in R^{4 \times 4}\}$, our goal is to predict the 3D bounding boxes $\{\textbf{b}_{j}\}$ and label scores $\{\textbf{s}_{j}\}$ of objects in the scene. Our pipeline borrows components from multi-view stereo (MVS)~\cite{murez2020atlas} and point cloud-based 3D detection methods~\cite{rukhovich2022fcaf3d}.
We first extract the $C$-channel 2D features ${\textbf{F}_{i} \in R^{h \times w \times C}}$ for each image with the 2D backbone $\mathcal{F}$. 
Then we aggregate the image features $\{\textbf{F}_{i}\}$ as volume feature $\textbf{V} \in R^{W \times H \times D \times C}$ with the assistance of associated camera parameters $\{\textbf{K}_{i}\}$ and $\{\textbf{R}_{i}\}$ using unprojection and average pooling~\cite{murez2020atlas, sun2021neuralrecon, rukhovich2022imvoxelnet}, which is denoted as $\mathcal{A}$.

Then we predict the rough scene TSDF $\textbf{S} \in R^{W \times H \times D}$ from the 3D volumes $\textbf{V}$ using the 3D reconstruction network $\mathcal{R}$ (Section~\ref{sec:3D reconstruction}). Section~\ref{sec:RMA} introduces a novel occlusion-aware aggregation module $\mathcal{A}^{*}$ to extract the 3D geometry as a point cloud with features $\textbf{P} \in R^{N_{pt} \times (3 + C)}$, based on the rough scene TSDF $\textbf{S}$ and a ray-marching-based voting scheme. Finally, the point cloud with features $\textbf{P}$ are passed through the detection network $\mathcal{D}$ to obtain the 3D bounding boxes $\{\textbf{b}_{j}\}$ and their corresponding label scores $\{\textbf{s}_{j}\}$ (Section~\ref{sec:3D detection network}).

\subsection{Multi-View Stereo Module}
\label{sec:3D reconstruction}
Complete reconstruction is important to avoid missing detection. While NeRF-based methods~\cite{mildenhall2021nerf, wang2021neus, yariv2021volume} deliver complete results, they require fitting model parameters for each specific scene. Since our task requires obtaining model parameters that can be universally applicable to all validation scenes, selecting them as our MVS module may lead to lower generalization ability or increased complexity in network training.
Among end-to-end 3D reconstruction methods, we find Atlas~\cite{murez2020atlas} a proper choice as our MVS module since it can be trained and used to predict the reconstruction in an end-to-end manner, including a 2D backbone and a 3D reconstruction network.

Specifically, for an input image $I_{i}$, we first extract the 2D features $\textbf{F}_{i}$ with $C$ channels using the ResNet50-FPN~\cite{lin2017feature} backbone. We then lift per-view 2D features into 3D via back projection given camera parameters $\{\textbf{K}_{i}\}$ and $\{\textbf{R}_{i}\}$, and aggregate them to generate 3D volume features $\textbf{V}$ with the voxel size of $4cm^{3}$ by average-pooling~\cite{murez2020atlas, sun2021neuralrecon, rukhovich2022imvoxelnet}. 

Next, we feed the 3D volume features into the 3D CNN reconstruction network in Atlas~\cite{murez2020atlas}, which features an encoder-decoder structure with skip connections~\cite{ronneberger2015u} with a $1\times1\times1$ convolutional head, to obtain the rough scene TSDF $\textbf{S}$. As suggested by \cite{murez2020atlas}, we employ the L1 loss at three different scales to boost the training:
\begin{equation}
L_{recon} = \sum_{i=1}^{3} \Vert \mathrm{sgn}(\textbf{S}_{i})\mathrm{ln}(1 + \lvert \textbf{S}_{i} \rvert) -  \mathrm{sgn}(\hat{\textbf{S}}_{i})\mathrm{ln}(1 + \lvert \hat{\textbf{S}}_{i} \rvert) \Vert_1
\label{recon_loss}
\end{equation}
where $\textbf{S}_{i}$ and $\hat{\textbf{S}}_{i}$ denotes the predicted and ground truth TSDF values from coarse to fine, and we denote the predicted TSDF with the finest scale $\textbf{S}_{3}$ as the rough scene TSDF $\textbf{S}$ utilized in the following sections.

\begin{figure}[t]
  \centering
   \includegraphics[width=1.0\linewidth]{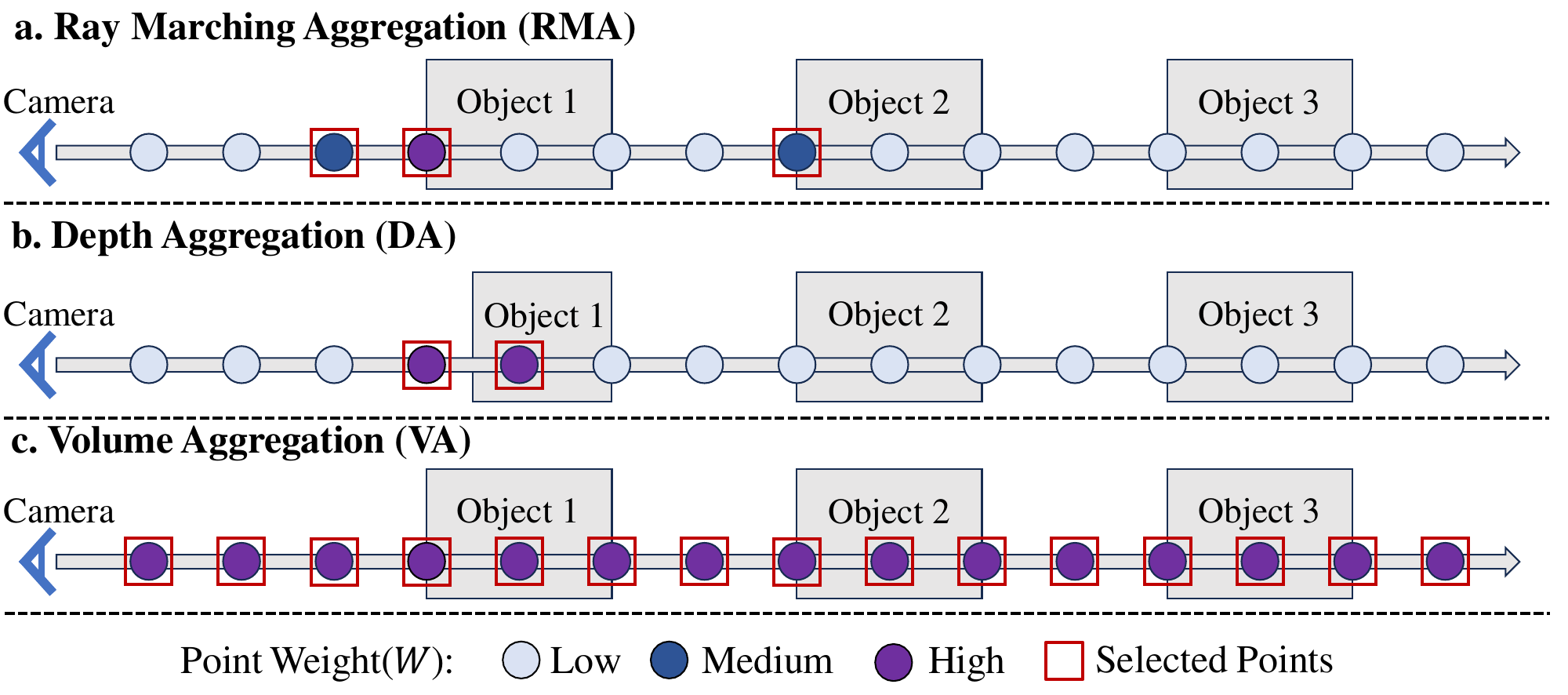}

   \caption{The 1D illustration comparing our Ray Marching Aggregation (RMA) method, with the Depth Aggregation method (DA) based on depth prediction, and the Volume Aggregation method(VA) based on unprojection~\cite{rukhovich2022imvoxelnet, sun2021neuralrecon, murez2020atlas}. The points depicted in the illustration represent sample points along a ray, with their colors indicating their respective weights. The points enclosed within one red square represent selected points.}
   \label{fig:rma}
\end{figure}

\subsection{Ray Marching Aggregation}
\label{sec:RMA}
Although we predict a 3D feature volume by directly averaging the lifted image features similar to \cite{rukhovich2022imvoxelnet, sun2021neuralrecon, murez2020atlas} in the reconstruction stage, volume features are polluted from certain views since image features can vote to unobserved space due to the lack of consideration for occlusion. An illustration of such aggregation from view to volume is illustrated in Figure~\ref{fig:rma}(c).

A straightforward solution to handle occlusion would be to directly render a depth map and vote image features only to the surface specified by the depth map (Figure~\ref{fig:rma}(b)). However, such a voting scheme is sensitive to the quality of the depth map produced from the reconstruction. In order to improve robustness, we introduce a soft occlusion-aware aggregation scheme called Ray Marching Aggregation (RMA), inspired by NeRF~\cite{mildenhall2021nerf} and NeuS~\cite{wang2021neus}.
Specifically, we compute the volume density given TSDF according to NeuS~\cite{wang2021neus}. We sample points on the ray of each pixel by ray marching and compute the opacity of each point by accumulating transmittance according to NeRF~\cite{mildenhall2021nerf}. As a result, we can compute 3D features by averaging image features weighted by the transmittance from different views. As illustrated in Figure~\ref{fig:rma}(a), RMA can vote image features into the scene by softly considering occlusions.
Finally, we extract points and aggregated features near the reconstructed surface and pass the point cloud to the 3D detection module for object detection.

In detail, given a 2D feature map $\textbf{F}_{i}$ and corresponding camera parameters $\textbf{K}_{i}$ and $\textbf{R}_{i}$, we assume that a ray $\textbf{p}_{i,u,v}(t) = \textbf{o}_{i} + t \cdot \textbf{d}_{i,u,v}$ is emitted from the camera towards the object represented by the pixel $\textbf{F}_{i}(u, v, :)$. For convenience, we denote the ray as $\textbf{p}(t) = \textbf{o} + t \cdot \textbf{d}$ in the following sections, where $\textbf{o}$ and $\textbf{d}$ can be determined as the origin and direction of the ray given the camera parameter and pixel location. We use ray marching to sample a set of points $\{\textbf{p}(t_i)\}$ along the ray with $t_i$ in increasing order.
Inspired by NeuS~\cite{wang2021neus}, the opacity value at the interval $[t_i,t_{i+1}]$ can be modeled as
\begin{equation}
\alpha(\textbf{p}(t_i)) = \max(\frac{\Phi(\textbf{S}(\textbf{p}(t_i))) - \Phi(\textbf{S}(\textbf{p}(t_{i+1})))}{\Phi(\textbf{S}(\textbf{p}(t_i)))}, 0),
\label{neus_alpha}
\end{equation}
with $\Phi(x)$ as the sigmoid function, and $\textbf{S}(\textbf{p}(t_i))$, which denotes the TSDF value of sample point $\textbf{p}(t_i)$, is obtained by querying the voxel closest to $\textbf{p}(t_i)$.
Then, the weight $W(\textbf{p}(t_i))$ of each sampled point $\textbf{p}(t_i)$ can be computed according to NeRF~\cite{mildenhall2021nerf} as
\begin{equation}
W(\textbf{p}(t_i)) = T(\textbf{p}(t_i)) \cdot \alpha(\textbf{p}(t_i)),
\label{neus_weight}
\end{equation}
where, $T(\textbf{p}(t_i))$ is the accumulated transmittance at the interval $[0,t_i]$ as
\begin{equation}
T(\textbf{p}(t_i)) = \prod \limits_{j=0}^{i-1}(1 - \alpha(\textbf{p}(t_j)))
\label{neus_T}
\end{equation}
We retain only those points with a weight greater than the threshold $\theta_{rma}$. 

To compute the 3D feature of each retained point, we average image features $\textbf{F}_{i}(u, v, :)$ weighted by the opacity of the point for $i$-th image, where $(u,v)$ is the projected location of the point to the image.
As a result, we obtain the point cloud with features $\textbf{P}$ by concatenating 3D coordinates and 3D features.

\subsection{3D Object Detection Network}
\label{sec:3D detection network}
We feed the reconstructed point cloud with aggregated feature $\textbf{P}$  into the 3D detection network $\mathcal{D}$ for the final detection results. We apply FCAF3D~\cite{rukhovich2022fcaf3d} as our detection network considering efficiency, memory consumption, and performance.

Firstly, we transform $\textbf{P}$ into sparse voxels~\cite{choy20194d} with a voxel size of $1cm^{3}$. Then we pass these sparse voxels into FCAF3D~\cite{rukhovich2022fcaf3d} to predict the classification scores $\textbf{s}$, bounding box regression parameters $\textbf{b}$, and 3D centerness $c$~\cite{tian2019fcos, rukhovich2022imvoxelnet, rukhovich2022fcaf3d} of each voxel. The detection loss $\mathcal{L}_D$ is the same as proposed in FCAF3D:
\begin{equation}L_{det} = \frac{1}{N_{pos}} \sum_{\hat{x}, \hat{y}, \hat{z}} (L_{cls}(\hat{\textbf{s}}, \textbf{s}) + m \cdot L_{reg}(\hat{\textbf{b}}, \textbf{b}) + m \cdot L_{cntr}(\hat{c}, c))\label{det_loss}\end{equation} Where $\{(\hat{x}, \hat{y}, \hat{z})\}$ represents the sparse voxel coordinates, $m = \vmathbb{1}_{\{\hat{\textbf{s}}_{\hat{x}, \hat{y}, \hat{z}} \neq 0\}}$ indicates whether a sparse voxel matches an object, $N_{pos} = \sum_{\hat{x}, \hat{y}, \hat{z}}m$ denotes the number of voxels matching an object. $L_{cls}$ is the focal loss to supervise $\textbf{s}$, $L_{reg}$ is the IOU loss for $\textbf{b}$, and $L_{cntr}$ is the binary cross-entropy loss for $c$.

\subsection{Training Procedure}
Due to the complexity of our architecture, which combines a MVS module and a detection network, training the modules from scratch may lead to overfitting. For example, to effectively train the detection network, it is essential to feed the network with high-quality point clouds with features, which makes the initialization of the reconstruction network important.  Therefore, we employ a pre-training and joint fine-tuning scheme in our training procedure to strike a balance between the 3D reconstruction network and the 3D detection network.

Firstly, we pre-train the 2D backbone and 3D reconstruction network using only the reconstruction loss $L_{recon}$, to fully leverage the 3D geometry. Subsequently, we freeze the aforementioned networks and proceed to pre-train the 3D detection network by utilizing the ray marching aggregation module and solely considering the detection loss $L_{det}$. Finally, to obtain the ultimate 3D detection results, we jointly fine-tune the entire network with the total loss denoted as $L_{t} = \lambda \cdot L_{recon} + L_{det}$. Where $\lambda$ is a constant to balance the reconstruction loss $L_{recon}$ and detection loss $L_{det}$.

%% file: sec/4_experiments.tex
\section{Experiments}
\label{sec:experiments}

\begin{figure*}[t]
  \centering
   \includegraphics[width=1\linewidth]{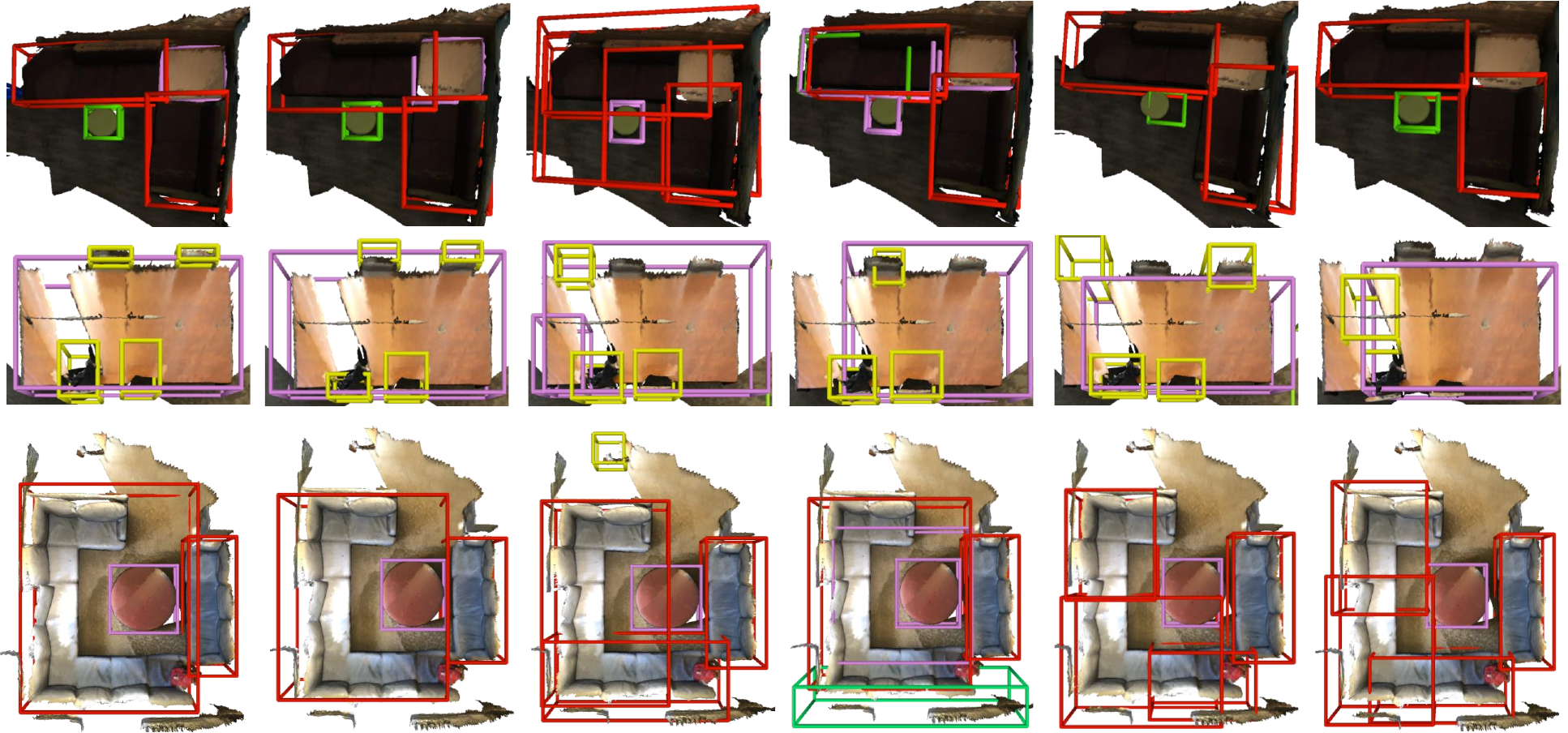}
   \flushleft{\qquad  Ground Truth \qquad \ \  CN-RMA \qquad  \ \ Atlas+FCAF  \qquad   \quad \ \ Neucon+FCAF  \qquad ImVoxelNet\cite{rukhovich2022imvoxelnet} \qquad NeRF-Det\cite{xu2023nerfdet}}\\
   \caption{\textbf{Visualization of 3D object detection results from ScanNet~\cite{dai2017scannet}.} From above to below are scene0559$\_$01, scene0598$\_$00, and scene0701$\_$00 from ScanNet. Atlas+FCAF denotes the two-stage baseline combining Atlas~\cite{murez2020atlas} and FCAF3D~\cite{rukhovich2022fcaf3d}, and Neucon+FCAF denotes the two-stage baseline combining NeuralRecon~\cite{sun2021neuralrecon} and FCAF3D. }
   \label{fig:vis_scannet_main}
\end{figure*}

\begin{figure*}[t]
  \centering
   \includegraphics[width=1\linewidth]{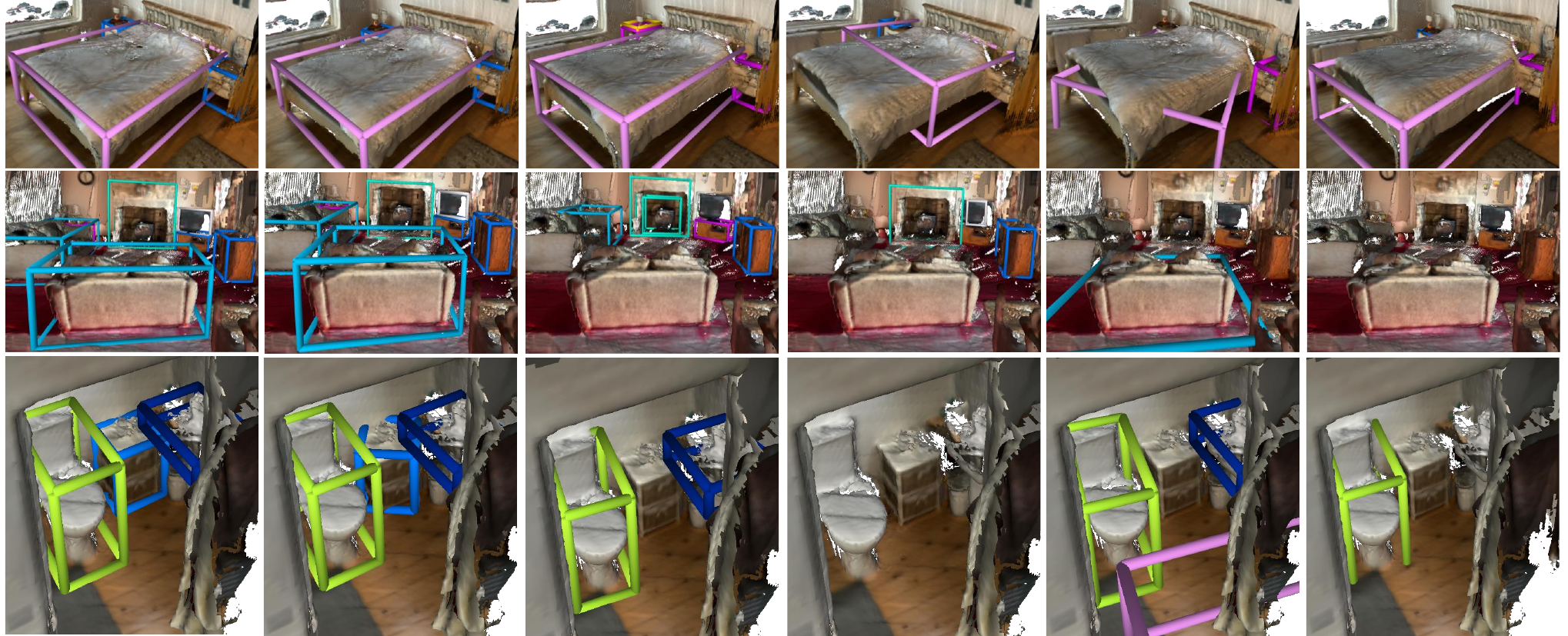}

   \flushleft{\qquad  Ground Truth \qquad \ \  CN-RMA \qquad  \ \ Atlas+FCAF  \qquad   \quad \ \ Neucon+FCAF  \qquad ImVoxelNet\cite{rukhovich2022imvoxelnet} \qquad NeRF-Det\cite{xu2023nerfdet}}\\
   \caption{\textbf{Visualization of 3D object detection results from ARKitScenes~\cite{dehghan2021arkitscenes}.} From above to below are scenes 44358583, 45663154, and 45261181 from ARKitScenes. Atlas+FCAF denotes the two-stage baseline combining Atlas~\cite{murez2020atlas} and FCAF3D~\cite{rukhovich2022fcaf3d}, and Neucon+FCAF denotes the two-stage baseline combining NeuralRecon~\cite{sun2021neuralrecon} and FCAF3D.}
   \label{fig:vis_arkit_main}
\end{figure*}

Section~\ref{Exp: datasets} introduces the datasets, metrics, and baselines in detail. The implementation details are introduced in Section~\ref{Exp: Implementation}. We compare our method with the state-of-the-art methods in Section~\ref{Exp: comparison}, where we show a clear advantage in terms of mAP@0.25 and mAP@0.5. Section~\ref{Exp: ablation} presents the ablation studies, and the improvements in the detection results demonstrate the effectiveness of our proposed ideas.

\subsection{Datasets, Metrics, and Baselines}
\label{Exp: datasets}
We evaluate our method, CN-RMA, using two indoor object detection datasets: ScanNet~\cite{dai2017scannet} and ARKitScenes~\cite{dehghan2021arkitscenes}. Following the settings of prior methods~\cite{rukhovich2022imvoxelnet, xu2023nerfdet, tu2023imgeonet}, we detect Axis-Aligned Bounding Boxes (AABB) for objects across 18 categories in ScanNet. The dataset is divided into 1201 training scans and 312 testing scans. For ARKitScenes, which comprises 4498 training scans and 549 testing scans, we detect Oriented Bounding Boxes (OBB) for objects across 17 categories. It is worth noting that the 3D geometric annotations in the ARKitScenes dataset are relatively rough. The depth map resolution of ARKitScenes is 192$\times$256, which is much lower compared to the 480$\times$640 resolution of ScanNet.
For the evaluation metrics, we choose the normally used mean average precision (mAP) with thresholds of 0.25 and 0.5 in both datasets.
For a fair comparison, we choose the previous state-of-the-art methods for indoor 3D object detection from multi-view images, namely ImVoxelNet~\cite{rukhovich2022imvoxelnet}, NeRF-Det~\cite{xu2023nerfdet}, and ImGeoNet~\cite{tu2023imgeonet}, as well as the two-stage baselines. As mentioned before, the two-stage baseline is a straightforward combination of 3D reconstruction and 3D detection methods. Specifically, Atlas~\cite{murez2020atlas} and NeuralRecon~\cite{sun2021neuralrecon} are utilized to reconstruct the 3D point clouds, while FCAF3D~\cite{rukhovich2022fcaf3d} is applied for the 3D detection.

\subsection{Implementation Details}
\label{Exp: Implementation}
CN-RMA is implemented using the MMDetection3D~\cite{mmdet3d2020} framework. We set the feature channels $C$ to 32. The weight threshold of our aggregation approach, $\theta_{rma}$, is set to 0.05. The loss weight $\lambda$ is set to 0.5. We sample 300 points for each pixel in ray marching, and the maximum $t$ is set as the diagonal length of the volume $\textbf{V}$. All experiments are conducted on 4 NVIDIA A6000 GPUs with a batch size of 1. More details are included in the supplementary material.

\begin{table}
  \centering
  \begin{tabular}{c|cc}
    \toprule
    Method&mAP@0.25$\uparrow$ & mAP@0.5$\uparrow$\\
    \midrule
    ImVoxelNet~\cite{rukhovich2022imvoxelnet} &46.7 &23.4 \\
    NeRF-Det~\cite{xu2023nerfdet} &53.5 &27.4 \\
    ImGeoNet~\cite{tu2023imgeonet} &54.8 &28.4 \\
Atlas~\cite{murez2020atlas}+FCAF3D~\cite{rukhovich2022fcaf3d}&55.4 &33.8 \\

NeuralRecon~\cite{sun2021neuralrecon}+FCAF3D &51.5 &31.6\\

    Ours (CN-RMA)&\textbf{58.6}&\textbf{36.8}\\
    \bottomrule
  \end{tabular}
    \caption{mAP@0.25 and mAP@0.5 results of the ScanNet~\cite{dai2017scannet} dataset. We directly cite the experimental results from the ImGeoNet~\cite{tu2023imgeonet} paper.}
      \label{tab:scannet}
\end{table}

\begin{table}
  \centering
  \begin{tabular}{c|cc}
    \toprule
    Method&mAP@0.25$\uparrow$&mAP@0.5$\uparrow$\\
    \midrule
    ImVoxelNet~\cite{rukhovich2022imvoxelnet}& 27.3 &4.3 \\
    NeRF-Det~\cite{xu2023nerfdet} &39.5 &21.9 \\
    ImGeoNet~\cite{tu2023imgeonet} &60.2 &43.4 \\
    Atlas~\cite{murez2020atlas}+FCAF3D~\cite{rukhovich2022fcaf3d} &51.3 &40.6 \\
    NeuralRecon~\cite{sun2021neuralrecon}+FCAF3D&36.3 &24.9\\

    Ours (CN-RMA)&\textbf{67.6} &\textbf{56.5}\\
    \bottomrule
  \end{tabular}
    \caption{mAP@0.25 and mAP@0.5 results of the ARKitScenes~\cite{dehghan2021arkitscenes} dataset. We directly cite the experimental results from the ImGeoNet~\cite{tu2023imgeonet} paper.}
      \label{tab:arkit}
\end{table}

\begin{table}
  \centering
  \begin{tabular}{c|c|cc}
    \toprule
    Method&parameter&mAP@0.25$\uparrow$&mAP@0.5$\uparrow$\\
    \midrule
    VA & $-$& 31.1 & 11.3\\
    RMA& $\theta_{rma}=0.02$ &\textbf{58.6} &\textbf{37.0} \\
    RMA& $\theta_{rma}=0.05$ &\textbf{58.6} &36.8 \\
    RMA& $\theta_{rma}=0.10$ &57.3 &35.4 \\
    DA& $k=1$ &57.1 &33.8 \\
    DA& $k=2$ &56.9 &34.1 \\
    DA& $k=3$ &56.3 &34.2 \\
    DA& $k=4$ &57.9 &34.7 \\
    \bottomrule
  \end{tabular}
  \caption{\textbf{Ablation study results of different aggregation schemes and parameters.} 
  VA refers to the Volume Aggregation method based on unprojection~\cite{rukhovich2022imvoxelnet, murez2020atlas, sun2021neuralrecon}. DA denotes the Depth Aggregation method relying on depth prediction. $\theta_{rma}$ represents the weight threshold for selecting sample points in our RMA method, while $k$ denotes the number of point pairs selected in the DA method. All experiments are conducted in ScanNet~\cite{dai2017scannet} with our proposed parameters following our standard training steps.}
  \label{tab:parameter}
\end{table}

\begin{table}
  \centering
  \begin{tabular}{c|cc}
    \toprule
    Training scheme&mAP@0.25$\uparrow$&mAP@0.5$\uparrow$\\
    \midrule
     Joint Train From Scratch&48.2&28.8 \\
     P-MVS + JFT&50.3&30.9\\
    P-MVS + P-Det&55.8&34.7\\
    P-MVS + P-Det + JFT&\textbf{58.6}&\textbf{36.8}\\
    \bottomrule
  \end{tabular}
  \caption{\textbf{Ablation study results of different training schemes.} P-MVS denotes pre-training the MVS module, P-Det denotes pre-training the detection network, and JFT denotes jointly fine-tuning the entire network. All experiments are conducted in ScanNet~\cite{dai2017scannet} with our proposed parameters.}
  \label{tab:train}
\end{table}

\subsection{Comparison}
\label{Exp: comparison}
We compare our method CN-RMA with the previous state-of-the-art method and two-stage baselines, as shown in Tables~\ref{tab:scannet} and~\ref{tab:arkit}. Our method achieves superior performance on both the ScanNet~\cite{dai2017scannet} and ARKitScenes~\cite{dehghan2021arkitscenes} datasets, outperforming other approaches in terms of mAP@0.25 and mAP@0.5. Our method surpasses the previous state-of-the-art method ImGeoNet~\cite{tu2023imgeonet} by 3.8 for mAP@0.25 and 8.4 for mAP@0.5 in ScanNet, and 7.4 for mAP@0.25 and 13.1 for mAP@0.5  in ARKitScenes. When compared to the two-stage baseline combining Atlas~\cite{murez2020atlas} and FCAF3D~\cite{rukhovich2022fcaf3d}, our method outperforms it by 3.2 for mAP@0.25 and 3.0 for mAP@0.5 in ScanNet, and 16.3 for mAP@0.25 and 15.9 for mAP@0.5 in ARKitScenes.
As shown in Figure~\ref{fig:vis_scannet_main}, ImVoxelNet and NeRF-Det often predict inaccurate bounding boxes due to insufficient utilization of geometric information. As for two-stage baselines, it is easy to predict inaccurate bounding boxes and miss some objects, due to noises and incomplete scene geometry reconstructed with MVS methods. 
Figure~\ref{fig:vis_arkit_main} shows the detection results of the ARKitScenes dataset. It is shown that the proposed method can also predict good results even with relatively low-quality reconstructed geometry, demonstrating the robustness of our method.
\subsection{Ablation Study}
\label{Exp: ablation}
In this section, we present the experimental results conducted on the ScanNet dataset to compare various aggregation schemes with different hyper-parameters and training schemes of our method.

\subsubsection{Aggregation Schemes}

We begin by comparing our occlusion-aware aggregation approach RMA, with the other two schemes shown in Figure~\ref{fig:rma}: Volume Aggregation (VA, Figure~\ref{fig:rma}(c)) and Depth Aggregation (DA, Figure~\ref{fig:rma}(b)). The VA method lifts per-view 2D features into 3D via back projection~\cite{sun2021neuralrecon, rukhovich2022imvoxelnet, murez2020atlas} directly. Specifically, we directly convert the global feature volume $\textbf{V}$ used in the 3D reconstruction network into a point cloud with features for detection. The DA method directly lifts 2D features to point clouds through depth maps of each view obtained from the reconstruction results.
In detail, for points along a ray, we define $\textbf{p}(t_{i})$ as the First Hitting Point (FHP) if
\begin{equation}
i = \mathop{\arg\min}\limits_{j}\{j \mid \textbf{S}(\textbf{p}(t_{j})) \cdot \textbf{S}(\textbf{p}(t_{j+1})) \leq 0\}
\label{da}
\end{equation}
which represents the first intersecting point. Considering the possible errors in 3D reconstruction, we select $2\cdot k$ points from $\textbf{p}(t_{i-k+1})$ to $\textbf{p}(t_{i+k})$. The weight of a selected point decreases linearly as it moves farther away from the FHP.

To study the sensitivity of the hyper-parameters, we explore different values of $\theta_{rma}$ in our RMA module with 0.02, 0.05, and 0.10. We refrain from testing smaller $\theta_{rma}$ values due to excessive GPU memory usage. For DA, we experiment with different values of k, including 1, 2, 3, and 4. 

Table~\ref{tab:parameter} presents the results of the different aggregation schemes and hyper-parameters. Our RMA method achieves the best performance in both mAP@0.25 and mAP@0.5, surpassing the VA method significantly by 27.5 in mAP@0.25 and 25.7 in mAP@0.5. The comparison reveals that integrating the rough scene TSDF obtained from 3D reconstruction into the aggregation process effectively enhances detection performance by providing valuable 3D geometry information and considering occlusion. Additionally, our RMA method outperforms the best results of DA by 0.7 in mAP@0.25 and 2.3 in mAP@0.5. It indicates that with the possible errors in the reconstructed scene TSDF, the DA method that directly chooses the first intersecting point along a ray may not always yield optimal results. In contrast, our RMA method offers more flexibility by combining local geometry information conveyed by $\alpha$ and accumulated ray information conveyed by $T$. Moreover, the results do not change much with different $\theta_{rma}$ indicating the robustness of our RMA method.

\subsubsection{Training Schemes}

We compare our training scheme, which involves subsequent pre-training of the MVS module and the detection network followed by joint fine-tuning of the entire network, with three other schemes to demonstrate the effectiveness of our proposed approach. The straightforward training scheme is to train the entire network from scratch without any pre-training. There are also several possible schemes considering the pre-training and fine-tuning. Specifically, the second scheme that we compare focuses on pre-training only the MVS module and then jointly training the entire network. The last scheme involves pre-training the MVS module and then freezing it to pre-train the detection module, without joint fine-tuning of the entire network.

The comparison results presented in Table~\ref{tab:train} demonstrate that our three-step training scheme with pre-training and fine-tuning achieves the best performance in both mAP@0.25 and mAP@0.5. Our training scheme outperforms the scheme without any pre-training by 10.4 in mAP@0.25 and 8.0 in mAP@0.5. Additionally, it outperforms the scheme without pre-training of the detection network by 8.3 in mAP@0.25 and 5.9 in mAP@0.5. These comparisons highlight the importance of both pre-training the MVS module and pre-training the reconstruction network for optimal performance. This is necessary to avoid potential overfitting caused by the complexity of our architecture and to provide a solid geometry foundation for our RMA aggregation scheme, which heavily relies on reliable geometry information from the reconstructed scene TSDF. Furthermore, our training scheme outperforms the scheme without fine-tuning by 2.8 in mAP@0.25 and 2.1 in mAP@0.5, demonstrating the effectiveness of fine-tuning. Fine-tuning facilitates knowledge transfer and synergistic interaction between the MVS module and the detection network, contributing to improved performance.

Overall, our experimental results validate the efficacy of our training scheme, emphasizing the significance of pre-training, fine-tuning, and the interplay between the MVS module and the detection network in achieving superior performance for our method.

%% file: sec/5_conclusion.tex
\section{Conclusion}
\label{sec:conclusion}
In this paper, we introduced CN-RMA, a novel 3D indoor object detection method from multi-view images. Our proposed approach surpasses previous state-of-the-art methods and outperforms two-stage baselines. We also present an effective occlusion-aware technique for aggregating 2D features into 3D point clouds using rough scene TSDF, which holds potential for integration into other 3D scene understanding tasks from multi-view images. 

Future work should focus on exploring techniques for further improving the performance of CN-RMA, such as investigating alternative aggregation schemes or incorporating additional contextual information, which could be beneficial. We anticipate continued advancements in 3D indoor object detection and related research areas by addressing these limitations and building upon our findings.

%% file: sec/X_suppl.tex
\clearpage
\setcounter{page}{1}
\maketitlesupplementary

\section{Implementation Details}

\subsection{Training Details}
In the pretraining stage of the Atlas~\cite{murez2020atlas} 2D backbone and reconstruction network (Stage 1), we directly utilize the checkpoint provided by Atlas to load our 2D backbone and 3D reconstruction network for ScanNet~\cite{dai2017scannet}. For ARKitScenes~\cite{dehghan2021arkitscenes}, we randomly select 50 images as input for each scene and set the voxel grid size to $(160\times160\times64)$. The network is trained for 80 epochs using the ADAM optimizer with a learning rate of 0.0005. In this stage, we perform data augmentation by applying random translations and rotations to the scenes following Atlas.

In the pretraining stage of the FCAF3D~\cite{rukhovich2022fcaf3d} detection network (Stage 2), we generate point clouds with features for each scene using our RMA method and use these point clouds to train the detection network. For ScanNet, we use 50 images and set the voxel grid size to $(256\times256\times96)$ for generating the point clouds. For ARKitScenes, we use 40 images and set the voxel grid size to $(192\times192\times80)$. We train the network for 12 epochs on both datasets, repeating the dataset 10 times for ScanNet and 3 times for ARKitScenes. The ADAM optimizer is employed with an initial learning rate of 0.001 and weight decay of 0.0001. The learning rate decreases on the 8th and 11th epochs. In this stage, we random sample $N_{pt}=500000$ points for each scene, and perform data augmentation by randomly translating, rotating, flipping, and scaling the input point clouds, following FCAF3D.

In the joint fine-tuning stage (Stage 3), we use 40 images and set the voxel grid size to $(192\times192\times80)$. The ADAM optimizer is used with an initial learning rate of 0.001 and weight decay of 0.0001. We fine-tune the ScanNet network for 100 epochs, with the learning rate decreasing on the 80th epoch. For ARKitScenes, we fine-tune the network for 40 epochs, with the learning rate decreasing on the 27th and 36th epochs. In this stage, we do not conduct random data augmentation on input camera parameters and ground truth TSDF. Instead, after obtaining point clouds with features, we random sample $N_{pt}=500000$ points for each scene, and perform data augmentation by randomly translating, rotating, flipping, and scaling the input point clouds, following FCAF3D.

For validation, we employ 50 images and set the grid size to $(256\times256\times96)$ for ScanNet, and we use 40 images and set the grid size to $(192\times192\times80)$ for ARKitScenes.

\subsection{Experiment Details}
Since the code has not been made publicly available, we directly reference the experimental results from the ImGeoNet~\cite{tu2023imgeonet} paper.

For experiments on ScanNet, we utilize the provided pre-trained networks of ImVoxelNet~\cite{rukhovich2022imvoxelnet}, NeRF-Det~\cite{xu2023nerfdet}, Atlas~\cite{murez2020atlas}, and NeuralRecon~\cite{sun2021neuralrecon}. For baselines with retraining, we retrain the FCAF3D~\cite{rukhovich2022fcaf3d} network for 12 epochs using the reconstructed scene point clouds, repeating the dataset 10 times.

For experiments on ARKitScenes, we train the Atlas network for 80 epochs and the NeuralRecon network for 15 epochs. We train the ImVoxelNet network for 12 epochs, repeating the dataset 9 times. We train the NeRF-Det network for 12 epochs without repeating the dataset. For baselines with retraining, we retrain the FCAF3D network for 12 epochs using the reconstructed scene point clouds, repeating the dataset 3 times.

\section{Additional Experiment Results}
\begin{figure*}[t]
  \centering
   \includegraphics[width=1\linewidth]{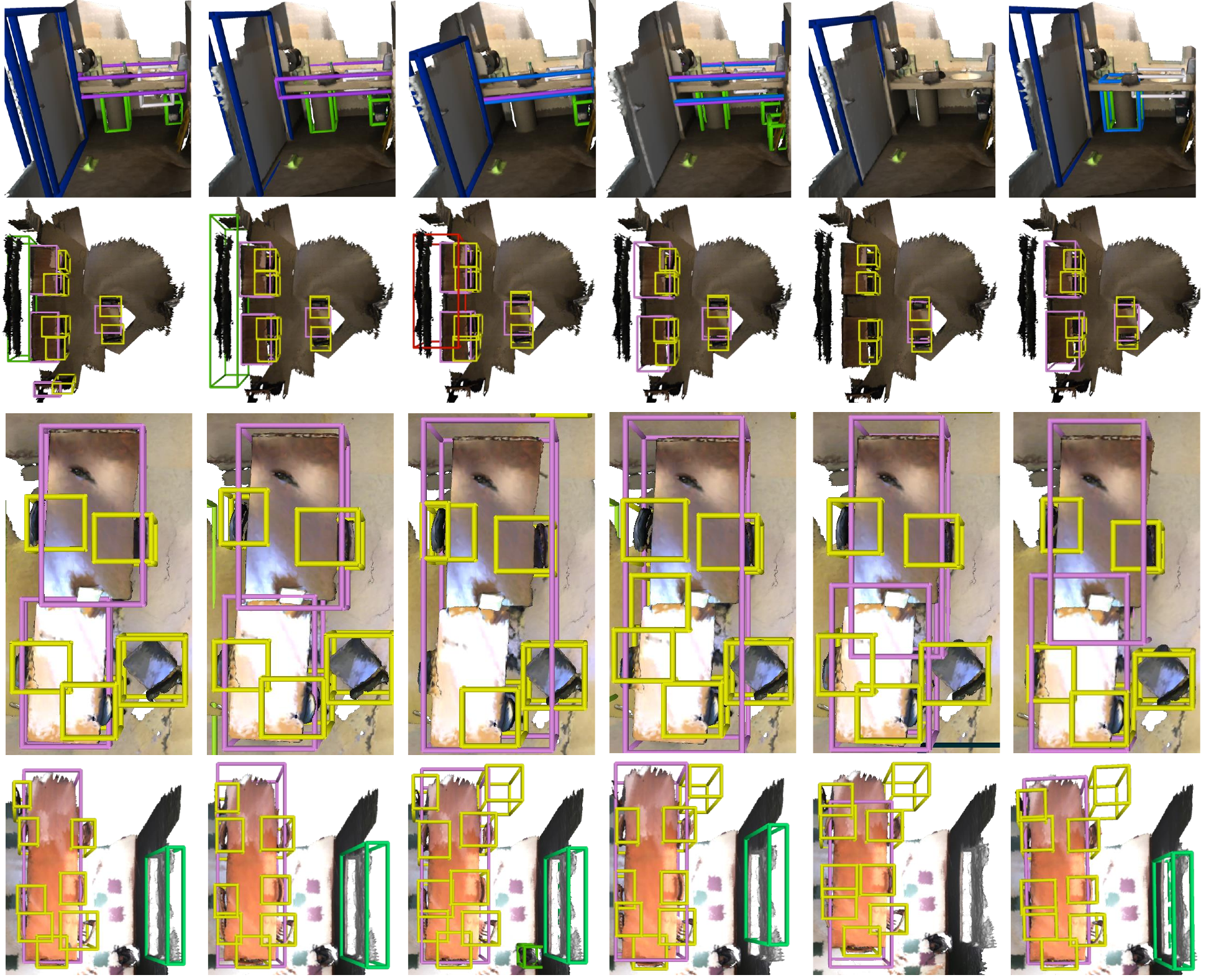}
   \flushleft{\qquad  Ground Truth \qquad \ \  CN-RMA \qquad  \ \ Atlas+FCAF  \qquad   \quad \ \ Neucon+FCAF  \qquad ImVoxelNet\cite{rukhovich2022imvoxelnet} \qquad NeRF-Det\cite{xu2023nerfdet}}\\
   \caption{\textbf{Visualization of 3D object detection results from ScanNet~\cite{dai2017scannet}.} From above to below are scene0084$\_$00, scene0609$\_$00, scene0030$\_$00, and scene0599$\_$02 from ScanNet. Atlas+FCAF denotes the two-stage baseline combining Atlas~\cite{murez2020atlas} and FCAF3D~\cite{rukhovich2022fcaf3d}, and Neucon+FCAF denotes the two-stage baseline combining NeuralRecon~\cite{sun2021neuralrecon} and FCAF3D. }
   \label{fig:vis_scannet_supp}
\end{figure*}

\begin{figure*}[t]
  \centering
   \includegraphics[width=1\linewidth]{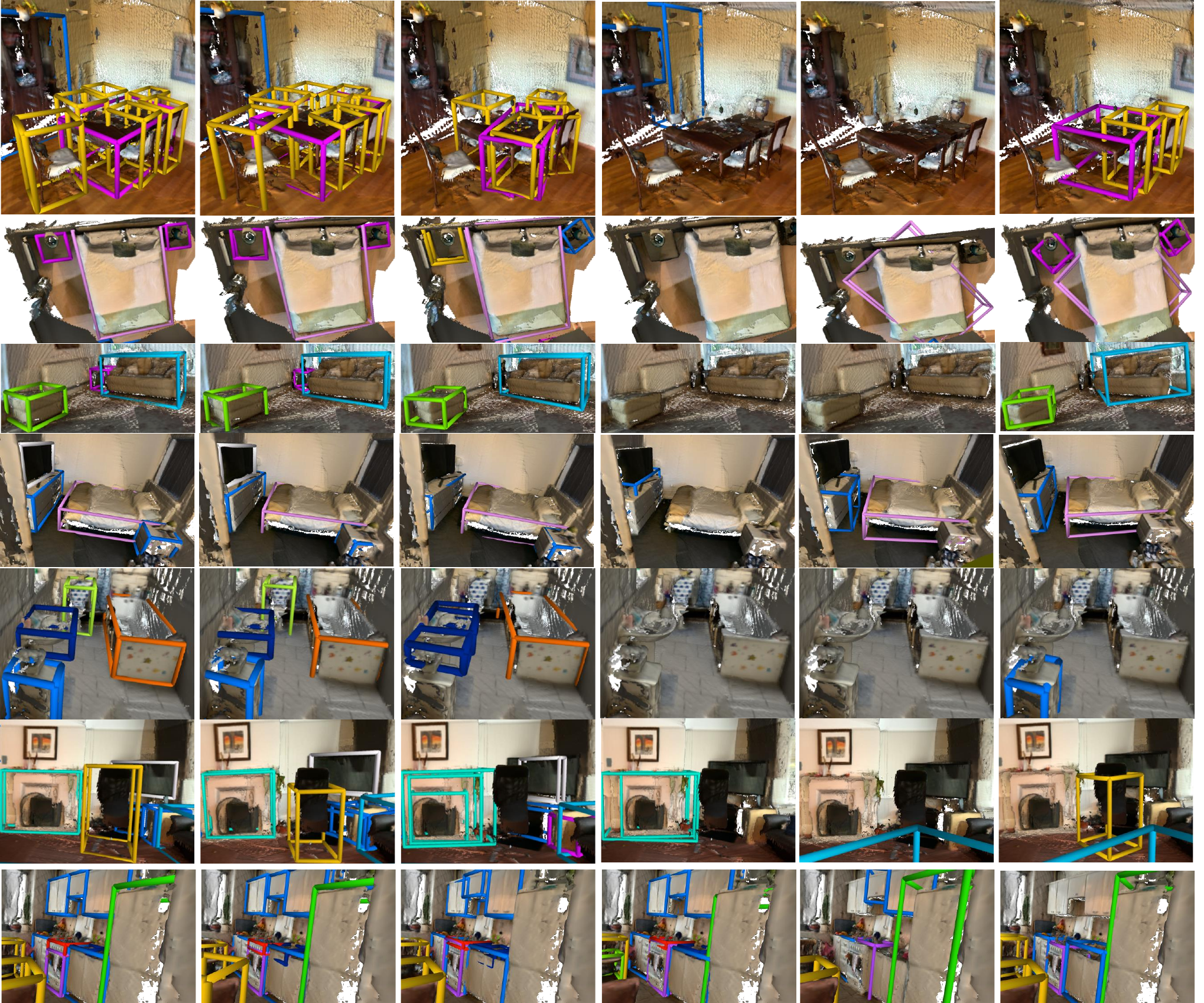}

   \flushleft{\qquad  Ground Truth \qquad \ \  CN-RMA \qquad  \ \ Atlas+FCAF  \qquad   \quad \ \ Neucon+FCAF  \qquad ImVoxelNet\cite{rukhovich2022imvoxelnet} \qquad NeRF-Det\cite{xu2023nerfdet}}\\
   \caption{\textbf{Visualization of 3D object detection results from ARKitScenes~\cite{dehghan2021arkitscenes}.} From above to below are scenes 47331266, 41069021, 42897538, 42445028, 45663114, 44358513, and 47333904 from ARKitScenes. Atlas+FCAF denotes the two-stage baseline combining Atlas~\cite{murez2020atlas} and FCAF3D~\cite{rukhovich2022fcaf3d}, and Neucon+FCAF denotes the two-stage baseline combining NeuralRecon~\cite{sun2021neuralrecon} and FCAF3D.}
   \label{fig:vis_arkit_supp}
\end{figure*}

The results of Table~\ref{tab:retrain} illustrate that retraining the reconstruction network of the two-stage baseline with reconstructed scene point clouds is better than directly using the network trained by ground truth point clouds. The per-category mAP@0.25 and mAP@0.5 scores of experiments on the ScanNet~\cite{dai2017scannet} and ARKitScenes~\cite{dehghan2021arkitscenes} datasets are shown in Tables~\ref{tab:scannet_25},~\ref{tab:scannet_5},~\ref{tab:arkit_25}, and~\ref{tab:arkit_5}. Additional visualization results of ScanNet and ARKitScenes are presented in Figures~\ref{fig:vis_scannet_supp} and ~\ref{fig:vis_arkit_supp}.

\begin{table*}
  \centering
  \begin{tabular}{c|c|c|cc}
    \toprule
    Method&Dataset&Retrain&mAP@0.25$\uparrow$ & mAP@0.5$\uparrow$\\
    \midrule
    Atlas~\cite{murez2020atlas}+FCAF3D~\cite{rukhovich2022fcaf3d} &ScanNet\cite{dai2017scannet}&\checkmark&55.4&33.8 \\
    Atlas+FCAF3D&ScanNet&&39.4&22.1\\
    NeuralRecon~\cite{sun2021neuralrecon}+FCAF3D &ScanNet&\checkmark&51.5&31.6 \\
    NeuralRecon+FCAF3D &ScanNet&&29.6&13.1\\
    Atlas+FCAF3D &ARKitScenes~\cite{dehghan2021arkitscenes}&\checkmark&51.3 &40.7\\
    Atlas+FCAF3D &ARKitScenes&&34.1&25.8\\
    NeuralRecon+FCAF3D &ARKitScenes&\checkmark&36.3&24.9\\
    NeuralRecon+FCAF3D &ARKitScenes&&13.6&9.1\\
    \bottomrule
  \end{tabular}
  \caption{Comparison of retraining for two-stage baselines.}
  \label{tab:retrain}
\end{table*}

\begin{table*}
  \centering
  \setlength{\tabcolsep}{0.65mm}
  \begin{tabular}{c|cccccccccccccccccc|c}
    \toprule
    Method&cab&bed&chair&sofa&tabl&door&wind&bkshf&pic&cntr&desk&curt&fridg&showr&toil&sink&bath&ofurn&mAP\\
    \midrule    ImVoxelNet~\cite{rukhovich2022imvoxelnet}&31.7&83.4&71.8&67.2&55.1&31.7&15.2&36.8&2.0&33.0&63.2&24.0&53.0&20.0&91.3&53.2&76.1&32.3&46.7 \\
    NeRF-Det~\cite{xu2023nerfdet}&\textbf{42.3}&84.6&75.9&78.5&56.3&33.4&21.4&49.9&2.4&50.6&\textbf{73.9}&21.3&\textbf{54.3}&\textbf{62.5}&90.9&57.7&75.5&32.3&53.5 \\
    ImGeoNet~\cite{tu2023imgeonet}&38.7&\textbf{86.5}&76.6&75.7&\textbf{59.3}&42.0&28.1&\textbf{59.2}&4.3&42.8&71.5&36.9&51.8&44.1&95.2&58.0&79.6&36.8&54.8\\
    Atlas+FCAF&41.6&85.4&\textbf{80.2}&81.6&54.7&38.3&27.3&50.1&7.6&58.9&73.3&16.8&36.6&61.9&94.4&58.8&\textbf{92.3}&37.4&55.4 \\
    Neucon+FCAF&38.7&82.2&78.3&81.4&56.2&30.5&12.5&42.1&6.0&54.2&64.6&20.8&34.6&41.3&89.1&67.8&89.1&37.3&51.5\\
    Ours(CN-RMA)&\textbf{42.3}&80.0&79.4&\textbf{83.1}&55.2&\textbf{44.0}&\textbf{30.6}&53.6&\textbf{8.8}&\textbf{65.0}&70.0&\textbf{44.9}&44.0&55.2&\textbf{95.4}&\textbf{68.1}&86.1&\textbf{49.7}&\textbf{58.6}\\
    
    \bottomrule
  \end{tabular}
  \caption{Per-category AP@0.25 scores for 18 categories from the ScanNet~\cite{dai2017scannet} dataset. Atlas+FCAF denotes the two-stage baseline combining Atlas~\cite{murez2020atlas} and FCAF3D~\cite{rukhovich2022fcaf3d}, and Neucon+FCAF denotes the two-stage baseline combining NeuralRecon~\cite{sun2021neuralrecon} and FCAF3D. We directly cite the experimental results from the ImGeoNet~\cite{tu2023imgeonet} paper.}
  \label{tab:scannet_25}
\end{table*}

\begin{table*}
  \centering
  \setlength{\tabcolsep}{0.65mm}
  \begin{tabular}{c|cccccccccccccccccc|c}
    \toprule
    Method&cab&bed&chair&sofa&tabl&door&wind&bkshf&pic&cntr&desk&curt&fridg&showr&toil&sink&bath&ofurn&mAP\\
    \midrule    ImVoxelNet~\cite{rukhovich2022imvoxelnet}&10.2&71.5&37.2&32.8&33.0&4.3&0.8&11.7&0.2&4.8&34.0&5.9&16.1&2.1&73.0&20.5&50.8&11.9&23.4 \\
    NeRF-Det~\cite{xu2023nerfdet}&15.8&73.1&45.3&40.6&39.5&8.1&2.0&20.3&0.2&13.8&42.5&5.3&25.3&\textbf{10.0}&63.0&26.0&49.1&12.7&27.4 \\
    ImGeoNet~\cite{tu2023imgeonet}&14.3&74.2&47.4&46.9&41.0&8.1&2.0&26.9&0.5&6.6&44.7&4.4&28.2&3.9&71.0&25.9&48.3&17.2&28.4\\
    Atlas+FCAF&20.3&\textbf{74.8}&47.9&65.0&\textbf{44.0}&9.7&5.0&37.0&1.1&\textbf{25.3}&\textbf{51.4}&3.5&23.4&3.0&69.3&31.8&74.1&22.6&33.8 \\
    Neucon+FCAF&15.8&74.7&45.6&\textbf{68.8}&43.2&8.0&3.5&26.1&\textbf{1.2}&15.8&40.4&1.3&21.9&1.6&\textbf{74.4}&28.3&\textbf{77.8}&21.3&31.6\\
    Ours(CN-RMA)&\textbf{21.3}&69.2&\textbf{52.4}&63.5&42.9&\textbf{11.1}&\textbf{6.5}&\textbf{40.0}&\textbf{1.2}&24.9&\textbf{51.4}&\textbf{19.6}&\textbf{33.0}&6.6&73.3&\textbf{36.1}&76.4&\textbf{31.5}&\textbf{36.8}\\
    \bottomrule
  \end{tabular}
  \caption{Per-category AP@0.5 scores for 18 categories from the ScanNet~\cite{dai2017scannet} dataset. Atlas+FCAF denotes the two-stage baseline combining Atlas~\cite{murez2020atlas} and FCAF3D~\cite{rukhovich2022fcaf3d}, and Neucon+FCAF denotes the two-stage baseline combining NeuralRecon~\cite{sun2021neuralrecon} and FCAF3D. We directly cite the experimental results from the ImGeoNet~\cite{tu2023imgeonet} paper.}
  \label{tab:scannet_5}
\end{table*}

\begin{table*}
  \centering
  \setlength{\tabcolsep}{0.65mm}
  \begin{tabular}{c|ccccccccccccccccc|c}
    \toprule
    Method&cab&fridg&shlf&stove&bed&sink&wshr&tolt&bthtb&oven&dshwshr&frplce&stool&chr&tble&TV&sofa&mAP\\
    \midrule    ImVoxelNet~\cite{rukhovich2022imvoxelnet}&20.7&33.3&13.5&4.3&57.7&25.8&53.8&65.2&66.0&25.5&2.2&0.2&2.5&26.7&24.5&0.0&41.6&27.3\\
    NeRF-Det~\cite{xu2023nerfdet}&34.7&61.1&30.7&9.4&73.2&29.9&62.6&77.2&86.4&45.0&7.4&2.1&12.1&46.4&38.3&0.1&55.5&39.5\\
    ImGeoNet~\cite{tu2023imgeonet}&55.8&\textbf{82.6}&\textbf{48.4}&20.4&89.3&52.8&\textbf{80.0}&92.5&94.7&66.0&18.1&\textbf{68.8}&30.6&72.3&70.3&2.2&79.0&60.2\\
    Atlas+FCAF&56.0&62.7&20.9&19.5&88.4&60.7&53.1&89.7&\textbf{94.9}&42.4&3.4&48.7&23.2&63.0&62.2&2.2&80.3&51.3 \\
    Neucon+FCAF&51.2&79.5&19.6&14.8&62.5&46.0&39.6&41.2&56.5&29.0&7.1&46.9&10.4&37.1&33.2&2.8&40.6&36.3 \\
    Ours(CN-RMA)&\textbf{73.5}&82.3&47.7&\textbf{37.4}&\textbf{91.6}&\textbf{74.5}&78.0&\textbf{93.3}&93.3&\textbf{74.6}&\textbf{53.6}&67.2&\textbf{35.9}&\textbf{73.1}&\textbf{72.8}&\textbf{14.5}&\textbf{85.1}&\textbf{67.6} \\
    \bottomrule
  \end{tabular}
  \caption{Per-category AP@0.25 scores for 17 categories from the ARKitScenes~\cite{dehghan2021arkitscenes} dataset. Atlas+FCAF denotes the two-stage baseline combining Atlas~\cite{murez2020atlas} and FCAF3D~\cite{rukhovich2022fcaf3d}, and Neucon+FCAF denotes the two-stage baseline combining NeuralRecon~\cite{sun2021neuralrecon} and FCAF3D. We directly cite the experimental results from the ImGeoNet~\cite{tu2023imgeonet} paper.}
  \label{tab:arkit_25}
\end{table*}

\begin{table*}
  \centering
  \setlength{\tabcolsep}{0.65mm}
  \begin{tabular}{c|ccccccccccccccccc|c}
    \toprule
    Method&cab&fridg&shlf&stove&bed&sink&wshr&tolt&bthtb&oven&dshwshr&frplce& stool&chr&tble&TV&sofa&mAP\\
    \midrule    ImVoxelNet~\cite{rukhovich2022imvoxelnet}&3.3&14.3&0.7&0.0&20.5&5.2&27.5&36.3&20.4&4.9&2.2&0.0&0.4&4.8&4.1&0.0&5.1&8.8\\
    NeRF-Det~\cite{xu2023nerfdet}&10.8&48.0&5.7&0.6&36.1&7.9&46.3&60.8&64.9&21.0&5.6&0.0&2.9&18.8&14.1&0.0&28.2&21.9 \\
    ImGeoNet~\cite{tu2023imgeonet}&31.8&72.5&21.7&3.9&83.3&19.9&71.2&84.8&91.0&44.4&15.9&23.1&13.3&49.3&45.1&0.1&67.2&43.4\\
    Atlas+FCAF&37.1&62.2&8.4&7.6&83.3&30.9&47.6&81.6&\textbf{94.1}&29.2&3.4&19.2&17.2&44.9&50.5&0.0&75.0&40.7 \\
    Neucon+FCAF&30.3&74.2&8.4&6.0&50.4&15.5&30.1&28.8&49.5&17.9&5.4&16.4&6.6&24.2&26.9&0.0&32.1&24.9 \\
    Ours(CN-RMA)&\textbf{56.0}&\textbf{79.8}&\textbf{27.8}&\textbf{21.3}&\textbf{87.4}&\textbf{51.6}&\textbf{75.9}&\textbf{89.1}&92.2&\textbf{60.8}&\textbf{53.3}&\textbf{40.6}&\textbf{25.1}&\textbf{60.5}&\textbf{60.1}&\textbf{1.2}&\textbf{77.3}&\textbf{56.5} \\
    \bottomrule
  \end{tabular}
  \caption{Per-category AP@0.5 scores for 17 categories from the ARKitScenes~\cite{dehghan2021arkitscenes} dataset. Atlas+FCAF denotes the two-stage baseline combining Atlas~\cite{murez2020atlas} and FCAF3D~\cite{rukhovich2022fcaf3d}, and Neucon+FCAF denotes the two-stage baseline combining NeuralRecon~\cite{sun2021neuralrecon} and FCAF3D. We directly cite the experimental results from the ImGeoNet~\cite{tu2023imgeonet} paper.}
  \label{tab:arkit_5}
\end{table*}

